\documentclass[default,sn-mathphys]{sn-jnl}

\usepackage{siunitx}

\usepackage{caption}
\usepackage{listings}
\usepackage{xcolor}

\definecolor{dkgreen}{rgb}{0,0.6,0}
\definecolor{gray}{rgb}{0.5,0.5,0.5}
\definecolor{mauve}{rgb}{0.58,0,0.82}

\DeclareCaptionLabelFormat{myformat}{\textbf{Algorithm~#2}}
\jyear{2023}%

\theoremstyle{thmstyleone}%
\theoremstyle{thmstyletwo}%

\theoremstyle{thmstylethree}%

\begin{document}

\title[Matters arising]{Limitations in odour recognition and generalisation in a neuromorphic olfactory circuit}

\author[1,2]{\fnm{Nik} \sur{Dennler}}\email{n.dennler2@herts.ac.uk}

\author[2]{\fnm{André} \sur{van Schaik}}\email{a.vanschaik@westernsydney.edu.au}

\author*[1]{\fnm{Michael} \sur{Schmuker}}\email{m.schmuker@herts.ac.uk}

\affil[1]{\orgdiv{Biocomputation Group}, \orgname{University of Hertfordshire}, \orgaddress{\street{College Lane}, \city{Hatfield}, \postcode{AL10 9AB}, \country{United Kingdom}}}

\affil[2]{\orgdiv{International Centre for Neuromorphic Systems, The MARCS Institute for Brain, Behaviour, and Development}, \orgname{Western Sydney University}, \orgaddress{\city{Kingswood}, \postcode{NSW 2747}, \country{Australia}}}

\abstract{
    Neuromorphic computing is one of the few current approaches that have the potential to significantly reduce power consumption in Machine Learning and Artificial Intelligence. Imam \& Cleland presented an odour-learning algorithm that runs on a neuromorphic architecture and is inspired by circuits described in the mammalian olfactory bulb. They assess the algorithm's performance in ``rapid online learning and identification'' of gaseous odorants and odorless gases (short ``gases'') using a set of gas sensor recordings of different odour presentations and corrupting them by impulse noise. We replicated parts of the study and discovered limitations that affect some of the conclusions drawn. First, the dataset used suffers from sensor drift and a non-randomised measurement protocol, rendering it of limited use for odour identification benchmarks. Second, we found that the model is restricted in its ability to generalise over repeated presentations of the same gas. We demonstrate that the task the study refers to can be solved with a simple hash table approach, matching or exceeding the reported results in accuracy and runtime. Therefore, a validation of the model that goes beyond restoring a learned data sample remains to be shown, in particular its suitability to odour identification tasks.
}

\maketitle

Imam \& Cleland's \cite{imam2020} algorithm takes inspiration from the neural pathways of the external plexiform layer of the mammalian olfactory bulb. 
Gas representations are built by an iterative approach of applying spike-time dependent plasticity rules to sequential gamma-frequency spike packages, on the basis of a dataset consisting of recordings from 72 Metal Oxide (MOx) gas sensors mounted in a wind tunnel \cite{vergara2013} (Fig. \ref{fig:fig1}a). 
They validate the model's capability to learn and robustly identify gases by computing and thresholding the Jaccard similarity coefficient between clean gases and representations arising from artificially occluded sensor recordings. 
Further aspects such as neuromodulation, contextual priming and neurogenesis are explored. 
The implementation and operation of the algorithm on Intel's Loihi neuromorphic platform \cite{Davies:2018} presents a major milestone in neuromorphic computing  due to the high complexity and biological realism of the underlying network model.
The authors claim to describe a gas identification framework that is superior to other models in terms of common classification metrics, that generalises ``broadly'' beyond experience and that can be deployed into environments containing unknown contaminants and other sources of interference \cite{imam2020}. 
In addition, the study has been referred to as a demonstration on how a neuromorphic network can learn and discriminate odours \cite{Davies2021,christensen20222022, DaviesNICE2021, Intel2020,Intel2020_1,Cornell2020}. 
Below we demonstrate limitations of the study that call these statements into question. 

\paragraph{}
The first limitation of the study relates to restrictions in the dataset used to
validate the olfactory bulb network. MOx sensors are highly prone to sensor drift, causing short- and long-term fluctuations in the sensors' baseline conductance and their responsiveness \cite{vergara2012}.
The most effective way of tackling the effect of drift on sensor response is to randomise the gas presentations over time during the recording. 
The dataset used here does not have this property: Recordings were acquired in gas-specific batches over the course of nine months (Fig. \ref{fig:fig1}b). 
The non-randomness, together with the dominating presence of sensor drift contaminations, allows for successful gas classification \textit{before} the gas is presented, which renders this dataset largely unsuitable for classification tasks \cite{dennler2022}. Drift contaminations could be partly mitigated by subtracting the baseline, i.e., the sensor response right before gas exposure \cite{hines1999electronic}. 
No baseline subtraction was performed in the discussed study, suggesting that the reported findings about odour learning and recognition may be skewed, and potentially invalid, due to the limitations of the dataset.

\begin{figure*}
    \includegraphics[width=\textwidth]{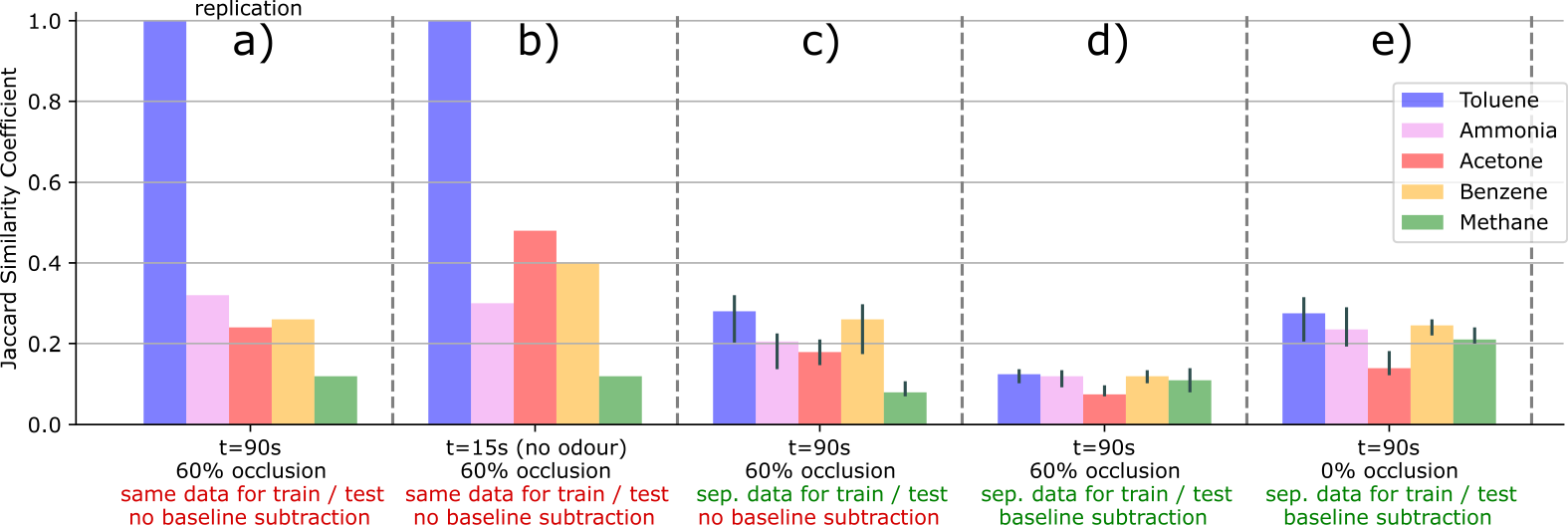}
    \caption{
    a) Jaccard similarity coefficient of the networks response to occluded Toluene and the learned odour representations, after five successive gamma cycles. Replicated from \cite{imam2020}. b) Spurious recognition in absence of gas in the wind tunnel. c,d,e) Recognition failure on different repetitions, c) without and d) with baseline subtraction, and e) without sample occlusion. 
    Median and interquartile range across 10 Toluene representations are displayed, and only five out of 10 gases are depicted for clarity.}
    \label{fig:fig2}
\end{figure*}

We repeated the simulations described by the authors for a range of conditions. 
As in the original work, the model was trained on 10 gases, and tested on 10 occlusions for each gas, i.e., 100 samples total. If not otherwise stated, 60\% of the data was occluded by impulse noise when testing.
We successfully replicated the authors' Jaccard similarity coefficient plot (\cite{imam2020}, Fig.\ 4b), using the same raw data points for composing training and test sets, and sampling the recordings at $t=90s$ (Fig. \ref{fig:fig1}a, \ref{fig:fig2}a). The result appears to demonstrate robust recognition of the Toluene gas instance.
Paradoxically, the same level of ``recognition'' of Toluene can be obtained in the absence of gas, using samples obtained at $t=15s$, \emph{before} the release of odour into the wind tunnel at $t=20s$ (Fig. \ref{fig:fig1}a, \ref{fig:fig2}b).
Therefore, the high Jaccard score for Toluene should be considered an artefact of sensor drift, and is unsuitable to substantiate a capability of the model to recognise odours \cite{dennler2022}.

\paragraph{}
In addition to the dataset's limitations, we found restrictions in the model's capability to generalise over different recordings of the same stimulus. 
Generalisation is an important property of any pattern recognition system \cite{gareth2013introduction}. 
The authors convincingly show that the model can restore input patterns corrupted by impulse noise.
However, in most instances the authors tested recognition on the same sample that was used for training, occluding 60\% of the sample with noise. 40\% of each training sample were present unchanged in the corresponding testing sample. A real odorant recognition and signal restoration system would rarely encounter the exact same stimulus twice, once in a clean and once in a corrupted version. Therefore, assessing the model's capability to recognise and restore patterns from separate recordings is essential to judge its relevance. 

For most gas and parameter combinations, the dataset contains 20 repetitions. We repeated the experiment 
using separate repetitions for training and testing and found that  

gas identity could not be recognised in occluded samples (Fig. \ref{fig:fig2}c).
Recognition scores were further reduced when subtracting the baseline from training and testing data (Fig. \ref{fig:fig2}d). 
In this configuration, aimed at mitigating sensor drift, recognition across repetitions failed even for samples without any noise occlusion (Fig. \ref{fig:fig2}e).

\begin{figure*}
    \includegraphics[width=\textwidth]{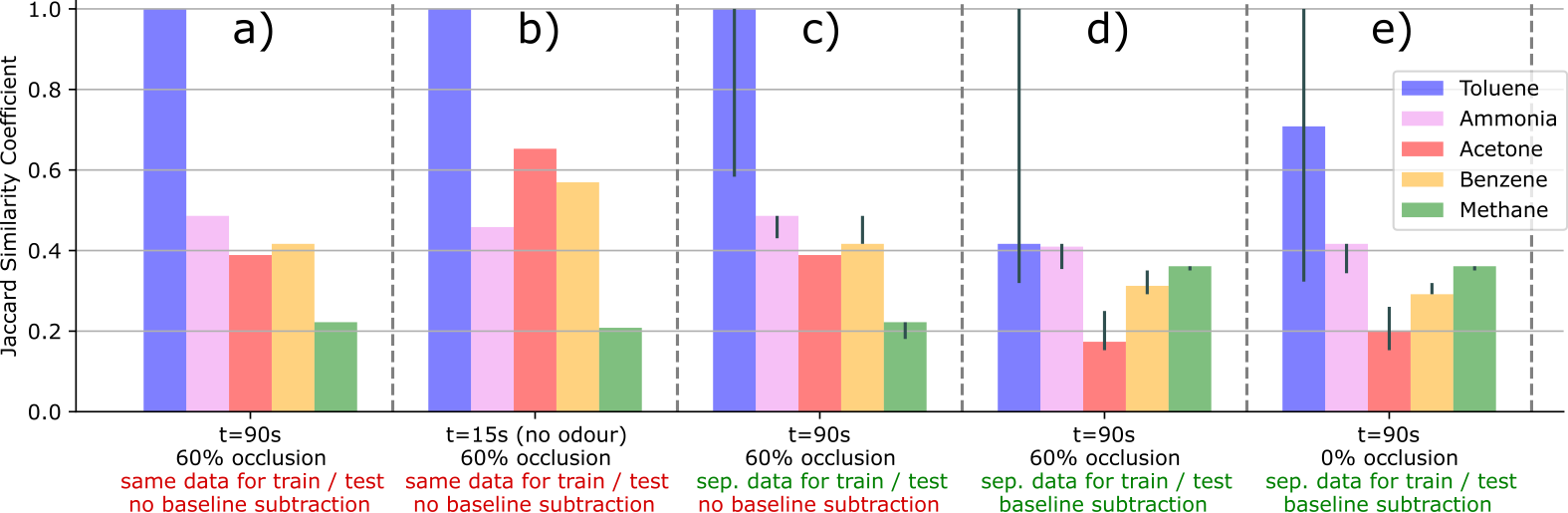}
    \caption{
    Jaccard similarity coefficient of the hash table denoiser's response to occluded Toluene and the learned odour representations. 
    Median and interquartile range across 10 Toluene representations are displayed, and only five out of 10 gases are depicted for clarity.}
    \label{fig:hashtable}
\end{figure*}

\paragraph{}
Finally, we demonstrate that the precise task undertaken in the study is trivial enough to be effectively addressed by using a simple hash table, such as a \textit{Python} dictionary, without the need for complex machine learning techniques. 
By storing the training samples in a hash table (i.e. one for each class, as in the original paper), and then computing and ranking the amount of overlap between a test sample and the stored training samples, one can estimate the most likely class (see Alg. \ref{algorithm:training}\&\ref{algorithm:testing}). 
Our implementation employs a concise one-shot approach, consisting of just 8 lines of \textit{Python} code, yet it matches or surpasses the suggested EPL network in recognition accuracy, and outperforms it in runtime (Fig.\ \ref{fig:hashtable}\&\ref{fig:execution_times}). In the light of these findings, the authors' claim of superior performance of the EPL network over advanced machine learning methods, derived from comparisons with multilayer denoising autoencoders and other methods,  cannot be upheld in its generality.

\paragraph{}
We conclude that the capability of the proposed model to identify learned odourants appears to be limited to corrupted versions of the training data. 
It failed to generalise to data outside the training set: Repetitions of learned gases were not recognised if these repetitions were not part of the training data.
Imam \& Cleland's model is an elegant example for an implementation of a biologically plausible model on neuromorphic hardware that can restore learned signals corrupted by noise. 
However, due to the restricted generalisation capability of the model, and to the limitations of the data used, it cannot be claimed that it solves the problem of odour learning and identification under a realistic scenario.
We hope that raising awareness about these limitations paves the way towards improved neuromorphic models for robust gas recognition that can solve real-world odour recognition tasks. 

\section*{Code and Data Availability}
Our experiments were based on the code released by the authors with the original study. Our adaptations and instructions for replication, together with the used data, are available at 
\url{https://github.com/BioMachineLearning/EPLNetwork_ImamCleland2020}.

\section*{Author Contribution Statement}
\textbf{Nik Dennler:} Conceptualisation, Investigation, Formal Analysis, Software, Visualisation, Writing – Original Draft, Writing – Review \& Editing. \textbf{André van Schaik:} Conceptualisation, Writing – Review \& Editing, Supervision. \textbf{Michael Schmuker:} Conceptualisation, Writing – Review \& Editing, Funding acquisition, Supervision. 

\section*{Declaration of Competing Interest}
The authors declare no competing interests.

\section*{Acknowledgements}
We thank N. Imam and T.A. Cleland for their valuable feedback and suggestions. Further, we thank D. Drix, M. Psarrou, S. Rastogi and S. Sutton for fruitful discussions.
M.S. was funded from EU H2020 Grant Human Brain Project
SGA3 (\#945539). This project is supported by the NSF/CIHR/DFG/FRQ/UKRI-MRC Next Generation Networks for Neuroscience Program (NSF \#2014217 / MRC \#MR/T046759/1 "Odor2Action").

\bibliography{sn-bibliography}


\begin{thebibliography}{13}
\ifx \bisbn   \undefined \def \bisbn  #1{ISBN #1}\fi
\ifx \binits  \undefined \def \binits#1{#1}\fi
\ifx \bauthor  \undefined \def \bauthor#1{#1}\fi
\ifx \batitle  \undefined \def \batitle#1{#1}\fi
\ifx \bjtitle  \undefined \def \bjtitle#1{#1}\fi
\ifx \bvolume  \undefined \def \bvolume#1{\textbf{#1}}\fi
\ifx \byear  \undefined \def \byear#1{#1}\fi
\ifx \bissue  \undefined \def \bissue#1{#1}\fi
\ifx \bfpage  \undefined \def \bfpage#1{#1}\fi
\ifx \blpage  \undefined \def \blpage #1{#1}\fi
\ifx \burl  \undefined \def \burl#1{\textsf{#1}}\fi
\ifx \doiurl  \undefined \def \doiurl#1{\url{https://doi.org/#1}}\fi
\ifx \betal  \undefined \def \betal{\textit{et al.}}\fi
\ifx \binstitute  \undefined \def \binstitute#1{#1}\fi
\ifx \binstitutionaled  \undefined \def \binstitutionaled#1{#1}\fi
\ifx \bctitle  \undefined \def \bctitle#1{#1}\fi
\ifx \beditor  \undefined \def \beditor#1{#1}\fi
\ifx \bpublisher  \undefined \def \bpublisher#1{#1}\fi
\ifx \bbtitle  \undefined \def \bbtitle#1{#1}\fi
\ifx \bedition  \undefined \def \bedition#1{#1}\fi
\ifx \bseriesno  \undefined \def \bseriesno#1{#1}\fi
\ifx \blocation  \undefined \def \blocation#1{#1}\fi
\ifx \bsertitle  \undefined \def \bsertitle#1{#1}\fi
\ifx \bsnm \undefined \def \bsnm#1{#1}\fi
\ifx \bsuffix \undefined \def \bsuffix#1{#1}\fi
\ifx \bparticle \undefined \def \bparticle#1{#1}\fi
\ifx \barticle \undefined \def \barticle#1{#1}\fi
\bibcommenthead
\ifx \bconfdate \undefined \def \bconfdate #1{#1}\fi
\ifx \botherref \undefined \def \botherref #1{#1}\fi
\ifx \url \undefined \def \url#1{\textsf{#1}}\fi
\ifx \bchapter \undefined \def \bchapter#1{#1}\fi
\ifx \bbook \undefined \def \bbook#1{#1}\fi
\ifx \bcomment \undefined \def \bcomment#1{#1}\fi
\ifx \oauthor \undefined \def \oauthor#1{#1}\fi
\ifx \citeauthoryear \undefined \def \citeauthoryear#1{#1}\fi
\ifx \endbibitem  \undefined \def \endbibitem {}\fi
\ifx \bconflocation  \undefined \def \bconflocation#1{#1}\fi
\ifx \arxivurl  \undefined \def \arxivurl#1{\textsf{#1}}\fi
\csname PreBibitemsHook\endcsname

\bibitem{imam2020}
\begin{barticle}
\bauthor{\bsnm{Imam}, \binits{N.}},
\bauthor{\bsnm{Cleland}, \binits{T.A.}}:
\batitle{Rapid online learning and robust recall in a neuromorphic olfactory
  circuit}.
\bjtitle{Nature Machine Intelligence}
\bvolume{2}(\bissue{3}),
\bfpage{181}--\blpage{191}
(\byear{2020}).
\doiurl{10.1038/s42256-020-0159-4}
\end{barticle}
\endbibitem

\bibitem{vergara2013}
\begin{barticle}
\bauthor{\bsnm{Vergara}, \binits{A.}},
\bauthor{\bsnm{Fonollosa}, \binits{J.}},
\bauthor{\bsnm{Mahiques}, \binits{J.}},
\bauthor{\bsnm{Trincavelli}, \binits{M.}},
\bauthor{\bsnm{Rulkov}, \binits{N.}},
\bauthor{\bsnm{Huerta}, \binits{R.}}:
\batitle{On the performance of gas sensor arrays in open sampling systems using
  inhibitory support vector machines}.
\bjtitle{Sensors and Actuators B: Chemical}
\bvolume{185},
\bfpage{462}--\blpage{477}
(\byear{2013}).
\doiurl{10.1016/j.snb.2013.05.027}
\end{barticle}
\endbibitem

\bibitem{Davies:2018}
\begin{barticle}
\bauthor{\bsnm{Davies}, \binits{M.}},
\bauthor{\bsnm{Srinivasa}, \binits{N.}},
\bauthor{\bsnm{Lin}, \binits{T.-H.}},
\bauthor{\bsnm{Chinya}, \binits{G.}},
\bauthor{\bsnm{Cao}, \binits{Y.}},
\bauthor{\bsnm{Choday}, \binits{S.H.}},
\bauthor{\bsnm{Dimou}, \binits{G.}},
\bauthor{\bsnm{Joshi}, \binits{P.}},
\bauthor{\bsnm{Imam}, \binits{N.}},
\bauthor{\bsnm{Jain}, \binits{S.}},
\bauthor{\bsnm{Liao}, \binits{Y.}},
\bauthor{\bsnm{Lin}, \binits{C.-K.}},
\bauthor{\bsnm{Lines}, \binits{A.}},
\bauthor{\bsnm{Liu}, \binits{R.}},
\bauthor{\bsnm{Mathaikutty}, \binits{D.}},
\bauthor{\bsnm{McCoy}, \binits{S.}},
\bauthor{\bsnm{Paul}, \binits{A.}},
\bauthor{\bsnm{Tse}, \binits{J.}},
\bauthor{\bsnm{Venkataramanan}, \binits{G.}},
\bauthor{\bsnm{Weng}, \binits{Y.-H.}},
\bauthor{\bsnm{Wild}, \binits{A.}},
\bauthor{\bsnm{Yang}, \binits{Y.}},
\bauthor{\bsnm{Wang}, \binits{H.}}:
\batitle{{Loihi: A Neuromorphic Manycore Processor with On-Chip Learning}}.
\bjtitle{IEEE Micro}
\bvolume{38}(\bissue{1}),
\bfpage{82}--\blpage{99}
(\byear{2018}).
\doiurl{10.1109/mm.2018.112130359}
\end{barticle}
\endbibitem

\bibitem{Davies2021}
\begin{barticle}
\bauthor{\bsnm{Davies}, \binits{M.}},
\bauthor{\bsnm{Wild}, \binits{A.}},
\bauthor{\bsnm{Orchard}, \binits{G.}},
\bauthor{\bsnm{Sandamirskaya}, \binits{Y.}},
\bauthor{\bsnm{Guerra}, \binits{G.A.F.}},
\bauthor{\bsnm{Joshi}, \binits{P.}},
\bauthor{\bsnm{Plank}, \binits{P.}},
\bauthor{\bsnm{Risbud}, \binits{S.R.}}:
\batitle{Advancing neuromorphic computing with loihi: A survey of results and
  outlook}.
\bjtitle{Proceedings of the IEEE}
\bvolume{109}(\bissue{5}),
\bfpage{911}--\blpage{934}
(\byear{2021}).
\doiurl{10.1109/JPROC.2021.3067593}
\end{barticle}
\endbibitem

\bibitem{christensen20222022}
\begin{barticle}
\bauthor{\bsnm{Christensen}, \binits{D.V.}},
\bauthor{\bsnm{Dittmann}, \binits{R.}},
\bauthor{\bsnm{Linares-Barranco}, \binits{B.}},
\bauthor{\bsnm{Sebastian}, \binits{A.}},
\bauthor{\bsnm{Le~Gallo}, \binits{M.}},
\bauthor{\bsnm{Redaelli}, \binits{A.}},
\bauthor{\bsnm{Slesazeck}, \binits{S.}},
\bauthor{\bsnm{Mikolajick}, \binits{T.}},
\bauthor{\bsnm{Spiga}, \binits{S.}},
\bauthor{\bsnm{Menzel}, \binits{S.}}, \betal:
\batitle{{2022 Roadmap on Neuromorphic Computing and Engineering}}.
\bjtitle{Neuromorphic Computing and Engineering}
\bvolume{2}(\bissue{2}),
\bfpage{022501}
(\byear{2022})
\end{barticle}
\endbibitem

\bibitem{DaviesNICE2021}
\begin{botherref}
\oauthor{\bsnm{Davies}, \binits{M.}}:
NICE 2021 Keynote: Lessons from Loihi for the Future of Neuromorphic Computing.
\url{https://www.youtube.com/watch?v=-dl1FPrpw1A&t=1463s}.
Accessed: 2022-11-22
(2021)
\end{botherref}
\endbibitem

\bibitem{Intel2020}
\begin{botherref}
\oauthor{\bsnm{Intel.Newsroom}}:
Computers That Smell: Intel’s Neuromorphic Chip Can Sniff Out Hazardous
  Chemicals.
\url{https://newsroom.intel.com/news/computers-smell-intels-neuromorphic-chip-sniff-hazardous-chemicals/}.
Accessed: 2022-10-24
\end{botherref}
\endbibitem

\bibitem{Intel2020_1}
\begin{botherref}
\oauthor{\bsnm{Intel.Newsroom}}:
How a Computer Chip Can Smell without a Nose.
\url{https://newsroom.intel.com/news/how-computer-chip-smell-without-nose/}.
Accessed: 2022-10-24
\end{botherref}
\endbibitem

\bibitem{Cornell2020}
\begin{botherref}
\oauthor{\bsnm{Lefkowitz}, \binits{M.}}:
Researchers sniff out AI breakthroughs in mammal brains.
\url{https://news.cornell.edu/stories/2020/03/researchers-sniff-out-ai-breakthroughs-mammal-brains}.
Accessed: 2022-11-22
(2020)
\end{botherref}
\endbibitem

\bibitem{vergara2012}
\begin{barticle}
\bauthor{\bsnm{Vergara}, \binits{A.}},
\bauthor{\bsnm{Vembu}, \binits{S.}},
\bauthor{\bsnm{Ayhan}, \binits{T.}},
\bauthor{\bsnm{Ryan}, \binits{M.A.}},
\bauthor{\bsnm{Homer}, \binits{M.L.}},
\bauthor{\bsnm{Huerta}, \binits{R.}}:
\batitle{Chemical gas sensor drift compensation using classifier ensembles}.
\bjtitle{Sensors and Actuators B: Chemical}
\bvolume{166-167},
\bfpage{320}--\blpage{329}
(\byear{2012}).
\doiurl{10.1016/j.snb.2012.01.074}
\end{barticle}
\endbibitem

\bibitem{dennler2022}
\begin{barticle}
\bauthor{\bsnm{Dennler}, \binits{N.}},
\bauthor{\bsnm{Rastogi}, \binits{S.}},
\bauthor{\bsnm{Fonollosa}, \binits{J.}},
\bauthor{\bsnm{{van Schaik}}, \binits{A.}},
\bauthor{\bsnm{Schmuker}, \binits{M.}}:
\batitle{Drift in a popular metal oxide sensor dataset reveals limitations for
  gas classification benchmarks}.
\bjtitle{Sensors and Actuators B: Chemical}
\bvolume{361},
\bfpage{131668}
(\byear{2022}).
\doiurl{10.1016/j.snb.2022.131668}
\end{barticle}
\endbibitem

\bibitem{hines1999electronic}
\begin{barticle}
\bauthor{\bsnm{Hines}, \binits{E.L.}},
\bauthor{\bsnm{Llobet}, \binits{E.}},
\bauthor{\bsnm{Gardner}, \binits{J.}}:
\batitle{Electronic noses: a review of signal processing techniques}.
\bjtitle{IEE Proceedings-Circuits, Devices and Systems}
\bvolume{146}(\bissue{6}),
\bfpage{297}--\blpage{310}
(\byear{1999})
\end{barticle}
\endbibitem

\bibitem{gareth2013introduction}
\begin{bbook}
\bauthor{\bsnm{Gareth}, \binits{J.}},
\bauthor{\bsnm{Daniela}, \binits{W.}},
\bauthor{\bsnm{Trevor}, \binits{H.}},
\bauthor{\bsnm{Robert}, \binits{T.}}:
\bbtitle{An Introduction to Statistical Learning: with Applications in R}.
\bpublisher{Springer},
\blocation{New York, NY}
(\byear{2013}).
\doiurl{10.1007/978-1-4614-7138-7}
\end{bbook}
\endbibitem

\end{thebibliography}

\newpage

\renewcommand{\thepage}{S\arabic{page}}
\renewcommand{\thesection}{S\arabic{section}}
\renewcommand{\thetable}{S\arabic{table}}
\renewcommand{\thefigure}{S\arabic{figure}}
\renewcommand{\figurename}{Figure}
\setcounter{figure}{0}

\section*{Supplementary Information}

\subsection*{Data Collection Protocol}
\begin{figure*}[h]
\centering
    \includegraphics[width=\linewidth]{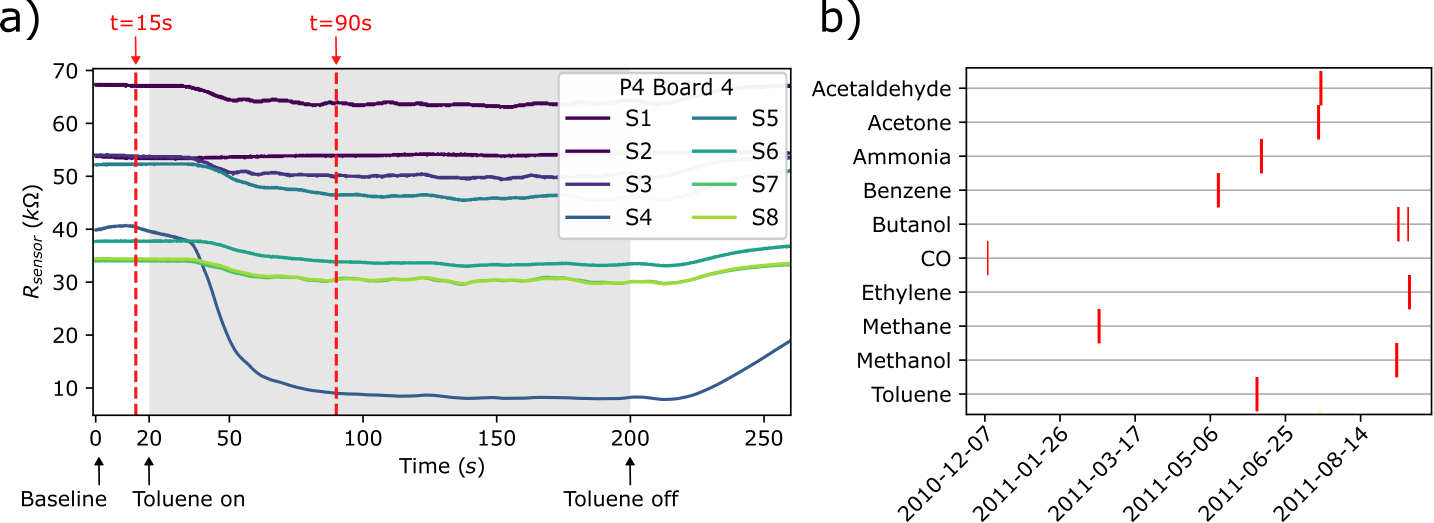}
    \caption{
    Overview of the data collection protocol.
    \textbf{a)} Example sensor resistance measurement for all sensors on one sensor board (location P4, module 4, Toluene, \SI{0.21}{\meter\per\second} airflow velocity, $\SI{5}{\volt}$ operating voltage, trial 2). The shaded area denotes the period during which the gas is injected into the wind tunnel, the red dotted lines indicate where the data is sampled in \cite{imam2020} ($t=\SI{90}{\second}$) and in this work ($t=\SI{15}{\second}$ and $t=\SI{90}{\second}$). \textbf{b)} Timestamps of the gas sensor recordings from which, in \cite{imam2020}, one trial per gas was sampled. Each vertical line represents multiple trials (up to 20), which were performed too close to each other for them to be visually distinguishable in this representation. Adapted from \cite{dennler2022}.}
\label{fig:fig1}.
\end{figure*}

\newpage
\subsection*{Hash Table Algorithm}

\begin{lstlisting}
learned_odour_representations = {idx: odour for idx, odour in enumerate(training_data)}
\end{lstlisting}
\begingroup
\captionof{lstlisting}{Training / One-shot learning (\textit{Python} code). For each training sample, fill dictionary with odour key and representation.}
\label{algorithm:training}
\endgroup

\begin{lstlisting}
denoised_testing_data = {}
for idx_test, testing_odour in enumerate(testing_data):
    best_matching, idx_best_matching = 0, None
    for idx_train, training_odour in learned_odour_representations.items():
        if best_matching < sum(training_odour==testing_odour):
            best_matching, idx_best_matching = sum(training_odour==testing_odour), idx_train
    denoised_testing_data[idx_test] = learned_odour_representations[idx_best_matching]
\end{lstlisting}
\begingroup
\captionof{lstlisting}{Testing / Denoising (\textit{Python} code). For each testing sample, compare how well the odour representation matches the training samples. Set denoised testing sample as the best-matching training sample.}
\label{algorithm:testing}
\endgroup

\begingroup
\begin{figure}[h]
    \centering
    \includegraphics[width=\linewidth]{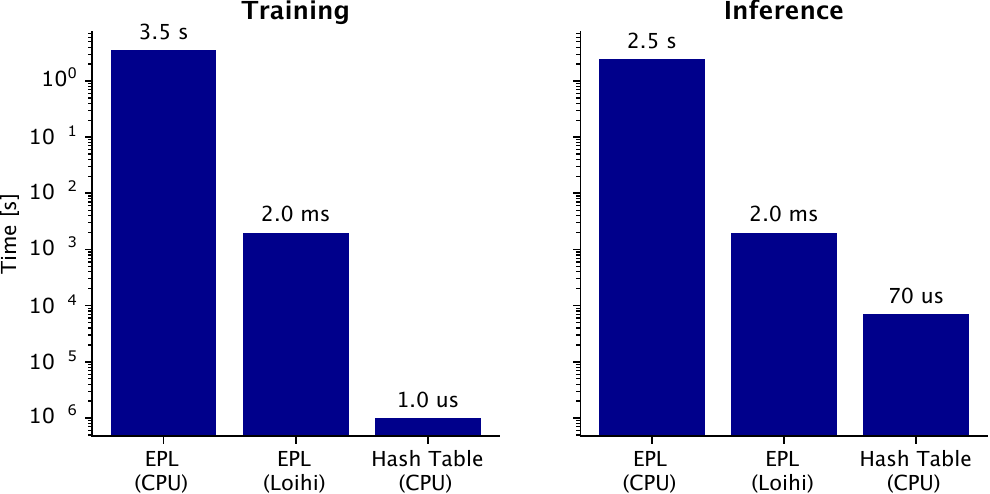}
\caption{Execution time comparison for the different algorithms and platforms. CPU execution times have been computed as the average of 10 and 100 training and inference instances respectively, run on a MacBook Pro M1. The Loihi execution time is denoted in the original paper \cite{imam2020}.}
\label{fig:execution_times}
\end{figure}
\endgroup

\end{document}